\documentclass[journal]{IEEEtran}

\ifCLASSINFOpdf
\usepackage[pdftex]{graphicx}
\graphicspath{{./}}

\DeclareGraphicsExtensions{.png,.pdf,.jpg,.jpeg}
\else
   \usepackage[dvips]{graphicx}
   \graphicspath{{Figures/}}
     \DeclareGraphicsExtensions{.png,.pdf,.jpg,.jpeg}
\fi

\usepackage{arydshln} 
\usepackage{fancyhdr}  

\usepackage{algorithmic}
\usepackage{algorithm}
\usepackage{array}
\usepackage[caption=false,font=normalsize,labelfont=sf,textfont=sf]{subfig}
\usepackage{textcomp}
\usepackage{stfloats}
\usepackage{url}
\usepackage{verbatim}
\usepackage{cite}
\usepackage{multirow}
\usepackage{pifont}
\usepackage{amsmath,amssymb,amsfonts,amsbsy}
\usepackage{bm} 
\usepackage{graphicx}
\usepackage{mathtools}
\usepackage{array,ragged2e}
\usepackage{booktabs}
\usepackage{makecell}

\usepackage{color}
\usepackage[table]{xcolor}
\definecolor{myPink}{rgb}{0.9294, 0.0078, 0.5490}
\definecolor{Gray}{gray}{0.92}
\definecolor{my_color}{HTML}{E8F3F1}
\definecolor{lightgray}{gray}{0.9}  
\usepackage[
  colorlinks=true,
  linkcolor=red,  
  citecolor=green,  
  urlcolor=myPink,   
  linkbordercolor={1 1 1} 
]{hyperref}
\usepackage{threeparttable}
\usepackage{diagbox}
\usepackage{orcidlink}

\newcommand{\cmark}{\textcolor{green!60!black}{\ding{51}}}
\newcommand{\xmark}{\textcolor{red}{\ding{55}}}

\makeatletter
\let\NAT@parse\undefined
\makeatother

\begin{document}
\title{Thresholded Cross-Attention for Reliable Intensity-Chromaticity Fusion in Low-Light Image Enhancement}

\author{
Yanyi~Wu$^*$,
Xu~Zhang$^*$,
Junkai~Chen,
Laibin~Chang,
Jiaqi~Ma,
Shi~Chen,
Linwei~Zhu,
Jianglei~Di,
and Huan~Zhang$^{\dagger}$%
\thanks{This work was supported by the National Natural Science Foundation of China under Grant 62302105, and in part by Funding by Science and Technology Projects in Guangzhou under Grant 2025A04J3851. \emph{(Yanyi Wu and Xu Zhang contributed equally to this work.)}\emph{(Corresponding author: Huan Zhang.)}}%
\thanks{Yanyi Wu, Junkai Chen, Jianglei Di, and Huan Zhang are with the School of Information Engineering, Guangdong University of Technology, Guangzhou 510000, China (e-mail: wuyanyi@mails.gdut.edu.cn; chenjunkai1@mails.gdut.edu.cn; jiangleidi@gdut.edu.cn; huanzhang2021@gdut.edu.cn).}%
\thanks{Xu Zhang and Laibin Chang are with the School of Computer Science, Wuhan University, Wuhan 430000, China (e-mail: zhangx0802@whu.edu.cn; changlb666@whu.edu.cn).}
\thanks{Jiaqi Ma is with the Department of Computer Vision, Mohamed bin Zayed University of Artificial Intelligence, Abu Dhabi, United Arab Emirates (e-mail: jiaqi.ma@mbzuai.ac.ae).}
\thanks{Shi Chen is with the Department of Computer Science, University of Macau, Macau 999078, China (e-mail: chenshi@um.edu.mo).}
\thanks{Linwei Zhu is with the Shenzhen Institutes of Advanced Technology, Chinese Academy of Sciences, Shenzhen 518055, China (e-mail: lw.zhu@siat.ac.cn).}
}

\markboth{Journal of \LaTeX\ Class Files,~Vol.~14, No.~8, August~2021}%
{Shell \MakeLowercase{\textit{\textit{et al.}}}: A Sample Article Using IEEEtran.cls for IEEE Journals}
\maketitle
\begin{abstract}

Low-Light Image Enhancement (LLIE) requires a careful balance among noise suppression, color fidelity, and efficiency. Recent HVI-based methods alleviate color entanglement by decoupling intensity and chromaticity, yet how reliably the two streams are fused again is an overlooked factor that largely determines the final quality. We observe that the confidence of cross-stream attention is strongly layer-dependent, so the fixed-quota selection of Top-K sparse attention is mismatched to it, discarding informative dependencies in some layers while retaining noisy ones in others.
Motivated by this observation, we propose TCA-Net, a network built around Thresholded Cross-Attention that targets reliable intensity--chromaticity fusion in the HVI space rather than introducing yet another color representation. At its core, TCA replaces the rigid Top-K quota with a fixed confidence threshold whose retained cardinality is input- and layer-adaptive, retaining only high-confidence cross-stream interactions while suppressing unreliable ones.
Around this core, two complementary designs clean up the fusion before and after it: a Phase-guided Fourier Interaction Module provides a structure-aware brightness initialization for the intensity stream prior to fusion, and a Decoupled Dual-Stream Guidance Module constructs residual intensity features to suppress chromaticity leakage during reconstruction. A Scale-Aware Consistency Regularization further improves structural robustness under scale perturbations during training. Extensive experiments on LOL-v1, LOL-v2, Sony-Total-Dark, and LSRW-Huawei demonstrate that TCA-Net delivers competitive restoration accuracy, improved color fidelity, and a compact parameter size. 
\end{abstract}

\IEEEpeerreviewmaketitle

\section{Introduction}

Low-light image enhancement (LLIE) aims to recover visually pleasing images with clear structures and faithful colors from images captured under insufficient illumination~\cite{li2021low}. Low-light images usually suffer from low contrast, noise amplification, color deviation, and loss of local details, which degrade both human visual perception and downstream vision tasks such as autonomous driving, video surveillance, and object detection~\cite{li2024light,lin2024dual,lu2024low}. Therefore, an effective LLIE method should not only improve brightness, but also suppress noise, preserve structural details, maintain color fidelity, and remain computationally efficient.

Early LLIE methods mainly rely on histogram equalization, gamma correction, or Retinex-based image formation models~\cite{6615961,6336819,jobson1997multiscale,guo2016lime,fu2016weighted}. Although these methods improve image visibility to some extent, they often suffer from over-enhancement, amplified noise, and unstable color restoration in complex real-world scenes. 
With the development of deep learning, data-driven methods have achieved remarkable progress, ranging from CNN- and Transformer-based enhancement networks to semi-supervised, contrastive, diffusion-, and prior-guided illumination restoration models~\cite{lore2017llnet,wei2018deep,guo2020zero,yang2020fidelity,liang2022semantically,InterLight,UniUIR,he2025reti,aaazhou2025low}. However, most of these methods operate directly in the RGB domain, where brightness, color, texture, and noise are highly entangled. As a result, enhancing brightness and restoring color simultaneously in RGB space may easily introduce color shifts, saturation distortion, and texture artifacts, especially under severe low-light degradation.

\begin{figure}[!t]
\centering
\includegraphics[width=\columnwidth]{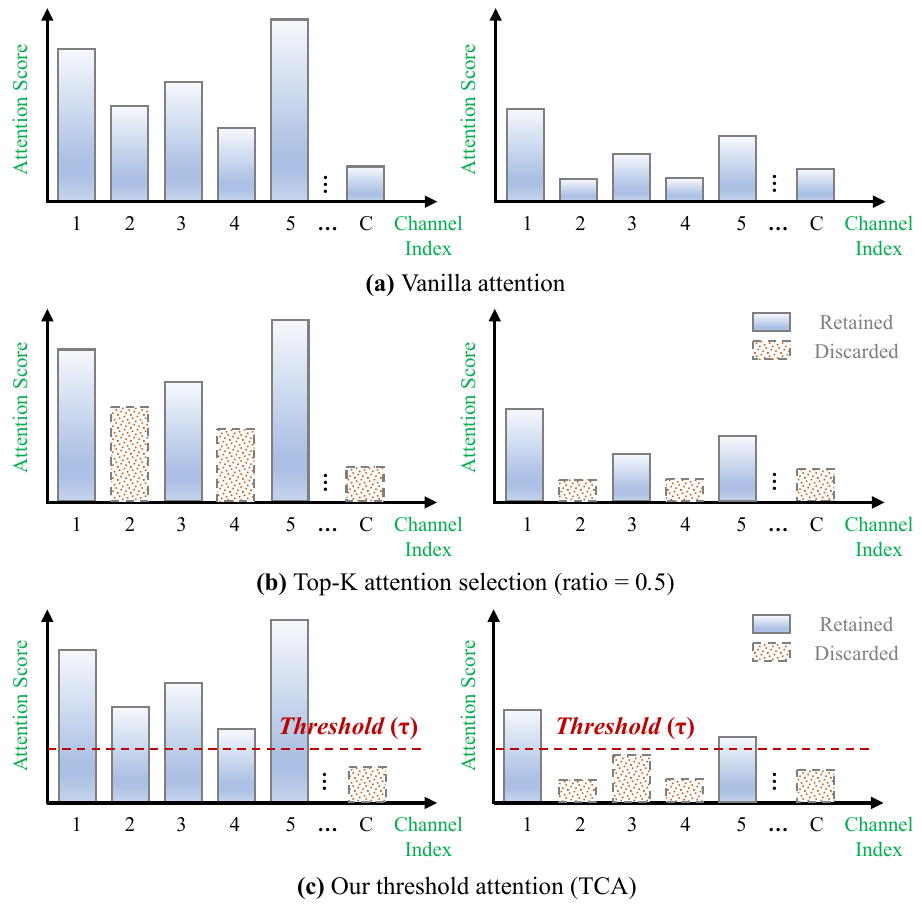}
\caption{
Motivation of the proposed Thresholded Cross-Attention (TCA), illustrated on channel-wise cross-stream attention. Each column shows a different attention distribution: a concentrated, high-confidence case (left) and a diffuse, low-confidence case (right). Solid blue bars are retained responses and dashed orange bars are discarded ones. (a) Vanilla attention keeps all channel responses, including weak noise-correlated ones. (b) Top-$K$ selection retains a fixed proportion ($\rho=0.5$) of the highest-ranking responses regardless of the distribution, so it discards informative responses in the concentrated case while being forced to keep weak ones in the diffuse case. (c) TCA instead applies a single confidence threshold $\tau$ shared across both cases: responses above $\tau$ are retained and those below are suppressed, so the number of retained responses adapts to the distribution, preserving confident dependencies while removing low-confidence ones.
}
\label{fig:motivation}
\end{figure}

To alleviate this issue, recent studies have explored color-consistent or color-space-based decomposition strategies. For example, color-consistency learning and YCbCr-based decomposition reduce color interference by separating luminance-related and chromatic components~\cite{00zhang2022deep,wang2024extracting}. More recently, the HVI color space has been introduced for LLIE to better decouple intensity and chromaticity, leading to improved color consistency and restoration stability~\cite{yan2025hvi}. Despite its effectiveness, HVI-based enhancement still faces an important limitation: after intensity and chromaticity are separated, reliable cross-stream interaction remains difficult. In severe low-light regions, the intensity stream may contain strong noise and unstable structural cues, while the chromaticity stream may include color artifacts and local distortions. Direct or dense cross-stream fusion can therefore propagate noise-corrupted intensity features to the chromaticity branch, or inject chromatic artifacts back into the reconstruction process.
 
\begin{figure}[!t]
    \centering
    \includegraphics[width=\columnwidth]{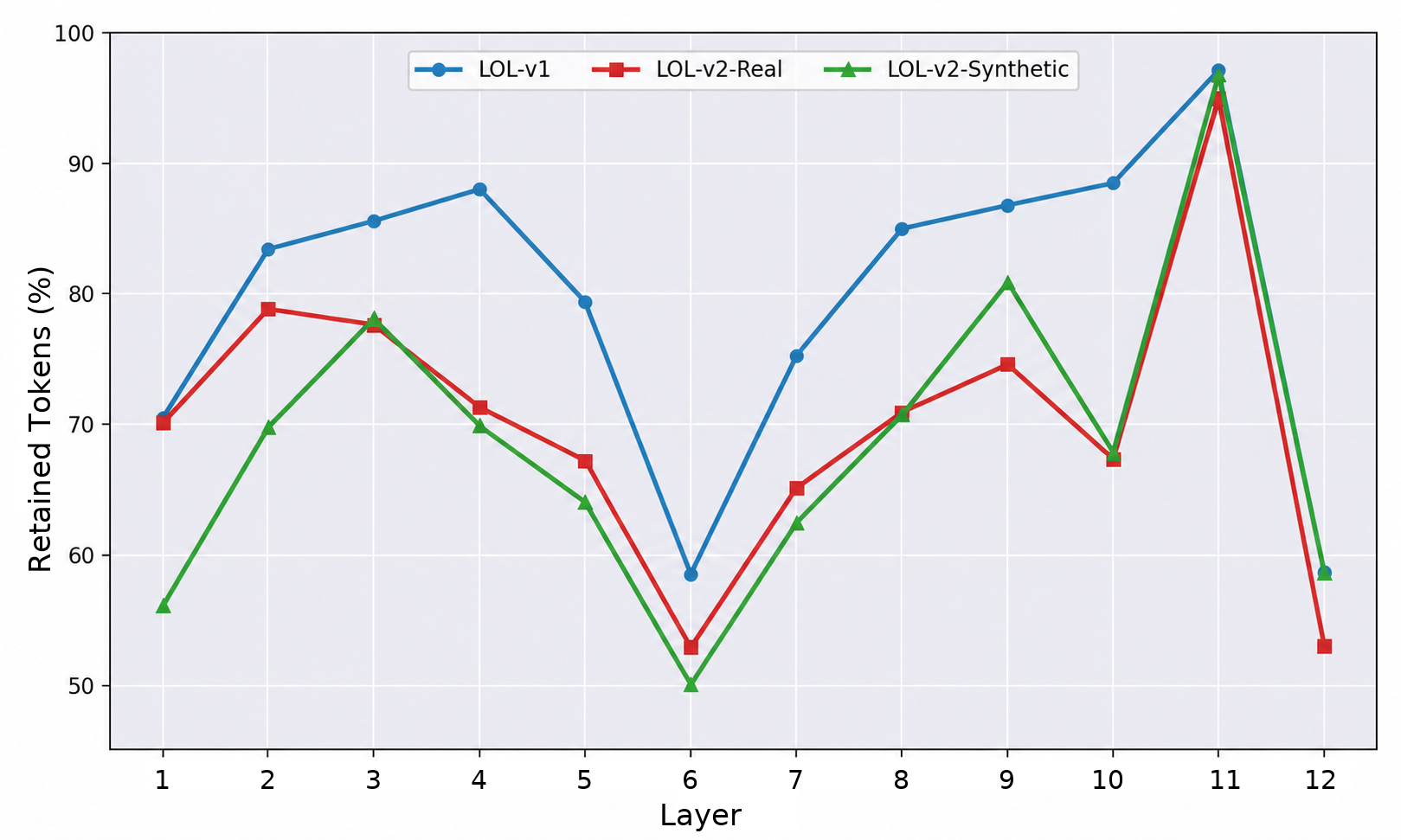}
\caption{Layer-wise retention rates of TCA on LOL-v1, LOL-v2-Real, and LOL-v2-Synthetic. Although the same threshold $\tau=0.025$ is used across layers, the retained ratio varies with the attention distribution of each layer and dataset. This shows that TCA produces adaptive sparse patterns without imposing a fixed retention quota.}
\label{fig:retention_rate}
\end{figure}

Existing attention-based fusion strategies are not fully suitable for this problem. Dense attention retains all cross-stream dependencies, including weak and noise-correlated responses. Top-K sparse attention reduces redundant interactions by preserving only a fixed proportion of high-ranking responses. However, we find that the reliability of cross-stream interactions is not uniform: the confidence of the cross-stream attention responses varies markedly across network depths, with some layers concentrating their weights in high-confidence ranges while others remain diffuse and low-valued. Against such a layer-dependent distribution, a fixed retention ratio is structurally mismatched. In concentrated layers it discards a fixed portion of informative cross-channel dependencies, while in diffuse layers it is forced to retain the same proportion of weak, noise-correlated responses simply to satisfy the quota. 
This behavior is illustrated conceptually in Fig.~\ref{fig:motivation} and further supported by the real cross-stream attention statistics in Fig.~\ref{fig:attn_hist}. Across different LOL benchmarks, the attention distributions vary clearly across layers: some layers concentrate more responses in high-confidence ranges, while others contain more diffuse low-confidence responses. Under such distributions, a fixed-ratio Top-$\mathrm{K}$ strategy may either discard informative dependencies or retain weak responses simply to satisfy the fixed quota. In contrast, threshold-based pruning selects interactions according to their confidence, allowing the retained ratio to adapt to the attention distribution. The corresponding layer-wise retention rates in Fig.~\ref{fig:retention_rate} further confirm that TCA does not impose a fixed sparse quota, but produces different retained ratios across encoder, bottleneck, and decoder layers. These observations motivate us to develop a reliability-aware fusion mechanism that preserves confident cross-stream interactions while suppressing unreliable ones.

Building upon this observation, we propose TCA-Net, a network for low-light image enhancement built around Thresholded Cross-Attention. Instead of introducing another color representation, TCA-Net is organized around a single objective: making the cross-stream fusion in the HVI space reliable. At the core of the framework is Thresholded Cross-Attention (TCA), which prunes cross-stream channel dependencies with a fixed confidence threshold rather than a fixed Top-K quota, so that the number of retained interactions adapts to the input- and layer-dependent attention distribution instead of a rigid cardinality. The remaining two designs serve this core by cleaning up the fusion before and after it. Before fusion, a Phase-guided Fourier Interaction Module (PFIM) uses phase-derived structural cues to recalibrate the amplitude response of intensity features, providing a structure-aware brightness initialization so that TCA operates on a less noise-corrupted intensity stream. After fusion, a Decoupled Dual-Stream Guidance Module (DDSGM) constructs residual intensity features by subtracting chromaticity-correlated components and progressively guides illumination and color restoration, thereby suppressing residual chromaticity leakage during reconstruction. Finally, a Scale-Aware Consistency Regularization (SACR) strategy is adopted during training to improve structural robustness under scale perturbations.
 
The main contributions are summarized as follows:
 
\begin{itemize}
    \item We identify reliability-aware cross-stream fusion as a key but overlooked bottleneck of HVI-based LLIE, and show that the confidence of cross-stream attention is layer-dependent and spatially non-uniform, which makes the fixed-quota selection of Top-K sparse attention systematically mismatched to it.
    \item Motivated by this observation, we propose TCA, which replaces the rigid Top-K quota with a fixed confidence threshold whose retained cardinality is input- and layer-adaptive, retaining high-confidence cross-stream dependencies and suppressing unreliable ones without the sorting overhead of rank-based selection.
 
    \item We complement TCA with a Phase-guided Fourier interaction module that delivers a structure-aware, less noise-corrupted intensity initialization prior to fusion, and a residual intensity-guided fusion module that suppresses chromaticity-correlated interference after fusion, so that the three designs jointly govern the cross-stream fusion from initialization to reconstruction.
    \item Extensive experiments on five LLIE benchmarks show that TCA-Net achieves competitive restoration performance across multiple evaluation settings, with improved color fidelity and a compact parameter size.
\end{itemize}

\section{Related work}
\subsection{Deep Learning-based Low-Light Image Enhancement}
Deep LLIE methods have evolved from direct low-to-normal mapping networks to more flexible supervised, unpaired, semi-supervised, and self-calibrated formulations~\cite{lv2018mbllen,jiang2021enlightengan,yang2020fidelity,ma2022toward}. Recent studies further address complex degradations and efficient representations, including joint enhancement and deblurring~\cite{lednet}, cross-data dehazing alignment~\cite{shi2025scaling}, context-conditioned implicit enhancement~\cite{chobola2024fast}, and state-space or RWKV-style restoration models~\cite{bai2024retinexmamba,di2024qmambabsr,xu2025urwkv}. Although these architectures improve restoration capacity, robustly balancing illumination recovery, noise suppression, and color fidelity remains challenging.

\begin{figure*}[!t]
    \centering
    \includegraphics[width=\textwidth]{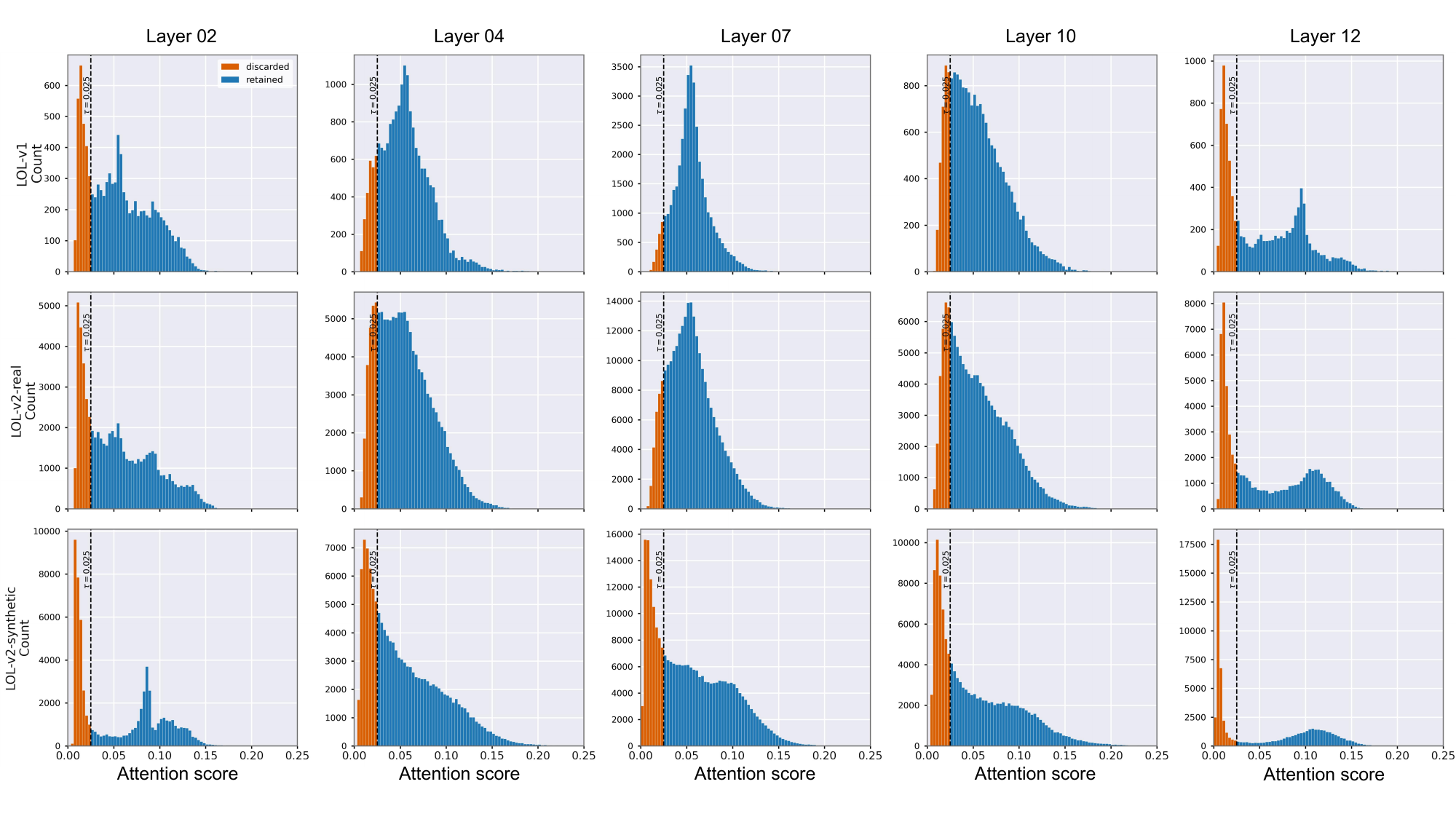}
\caption{Softmax-normalized cross-stream attention score distributions before selection. The three rows correspond to LOL-v1, LOL-v2-Real, and LOL-v2-Synthetic, respectively. The sampled layers follow the network depth: layers 1--4 are encoder layers, layers 5--8 are bottleneck layers, and layers 9--12 are decoder layers. Blue and orange bars denote responses retained and discarded by TCA with $\tau=0.025$. The layer- and dataset-dependent distributions show why a fixed retention quota is mismatched to cross-stream confidence.}
    \label{fig:attn_hist}
\end{figure*}

Recent LLIE studies also incorporate visual priors and real-world degradation modeling, including multi-prior collaboration~\cite{she2024mpcnet}, generative perceptual priors~\cite{aaazhou2025low}, cross-image disentanglement~\cite{guo2024cidn}, and darkness-aware visible--infrared fusion~\cite{zhou2025dfvo}. However, RGB-domain enhancement can introduce color shifts, amplify noise, and destabilize local details, since illumination, texture, and chromatic information are strongly entangled~\cite{wei2018deep,guo2020zero}.

To reduce this entanglement, decoupled enhancement frameworks separate illumination-related components from other visual factors. Representative examples include Retinex-based decomposition~\cite{wei2018deep,zhang2019kindling}, color-consistent or YCbCr-based modeling~\cite{00zhang2022deep,wang2024extracting}, HVI-space enhancement~\cite{yan2025hvi}, and Fourier-based restoration~\cite{wang2023fourllie,li2023embedding,zhang2024dmfourllie,she2025exploring}. These methods improve illumination and color restoration from different perspectives, but reliable interaction between intensity and chromaticity features after decomposition remains less explored. Different from existing HVI-based or frequency-based methods, TCA-Net focuses on reliability-aware intensity--chromaticity fusion through frequency-guided intensity initialization, confidence-thresholded cross-stream pruning, and residual intensity-guided reconstruction.

\subsection{Confidence-Driven Sparse Attention}

Sparse attention has been widely studied to reduce redundant interactions in Transformers. Early methods use fixed sparse patterns to lower memory and computational costs while preserving long-range modeling ability, as in Sparse Transformer~\cite{child2019generating} and BigBird~\cite{zaheer2020big}. Later methods introduce content-adaptive sparsity, where the attended regions or tokens are selected according to input features rather than a predefined pattern~\cite{zhu2020deformable,zhu2023biformer}. These designs show that attention sparsity should not only reduce redundancy, but also preserve informative dependencies under varying visual content.

In image restoration and enhancement, sparse attention is often introduced to handle high-resolution feature maps efficiently. Window-based attention, such as Swin Transformer~\cite{liu2021swin}, restricts interactions to local windows and therefore reduces redundant global computation. However, low-light enhancement often requires consistent illumination recovery across distant regions, making purely local sparse patterns less suitable. Dynamic selection methods provide a more flexible alternative. For example, restoration-specific models such as DRSformer~\cite{chen2023learning} use Top-$\mathrm{K}$ selection to retain the highest-ranking attention responses. Nevertheless, Top-$\mathrm{K}$ sparsification still relies on a manually fixed retention ratio, which may not match layer- or input-dependent attention distributions. It may discard useful responses in concentrated distributions or retain weak responses in diffuse ones.

Another group of methods induces sparsity by modifying the normalization function itself. Sparsemax~\cite{martins2016softmax} and Entmax~\cite{correia2019adaptively} produce sparse probability distributions through simplex projection, while Rectified Linear Attention~\cite{zhang2021sparse} replaces Softmax with a ReLU-based formulation. These methods can suppress low-scoring entries, but the pruning criterion is implicit in the projection or activation and may require additional sorting or iterative computation. For low-light enhancement, standard Softmax can also assign non-zero probabilities to noise-heavy responses~\cite{Bondarenko2023QuantizableTR}, which may propagate unreliable intensity or chromaticity cues. Motivated by these observations, we keep the standard Softmax attention and apply an explicit confidence threshold after normalization. This provides an interpretable probability-level pruning rule, avoids a fixed retention quota, and better supports reliable intensity--chromaticity fusion.
\section{Method}

\subsection{Overall Pipeline and Training Objective}

\begin{figure*}[!t]
\centering
\includegraphics[width=\textwidth]{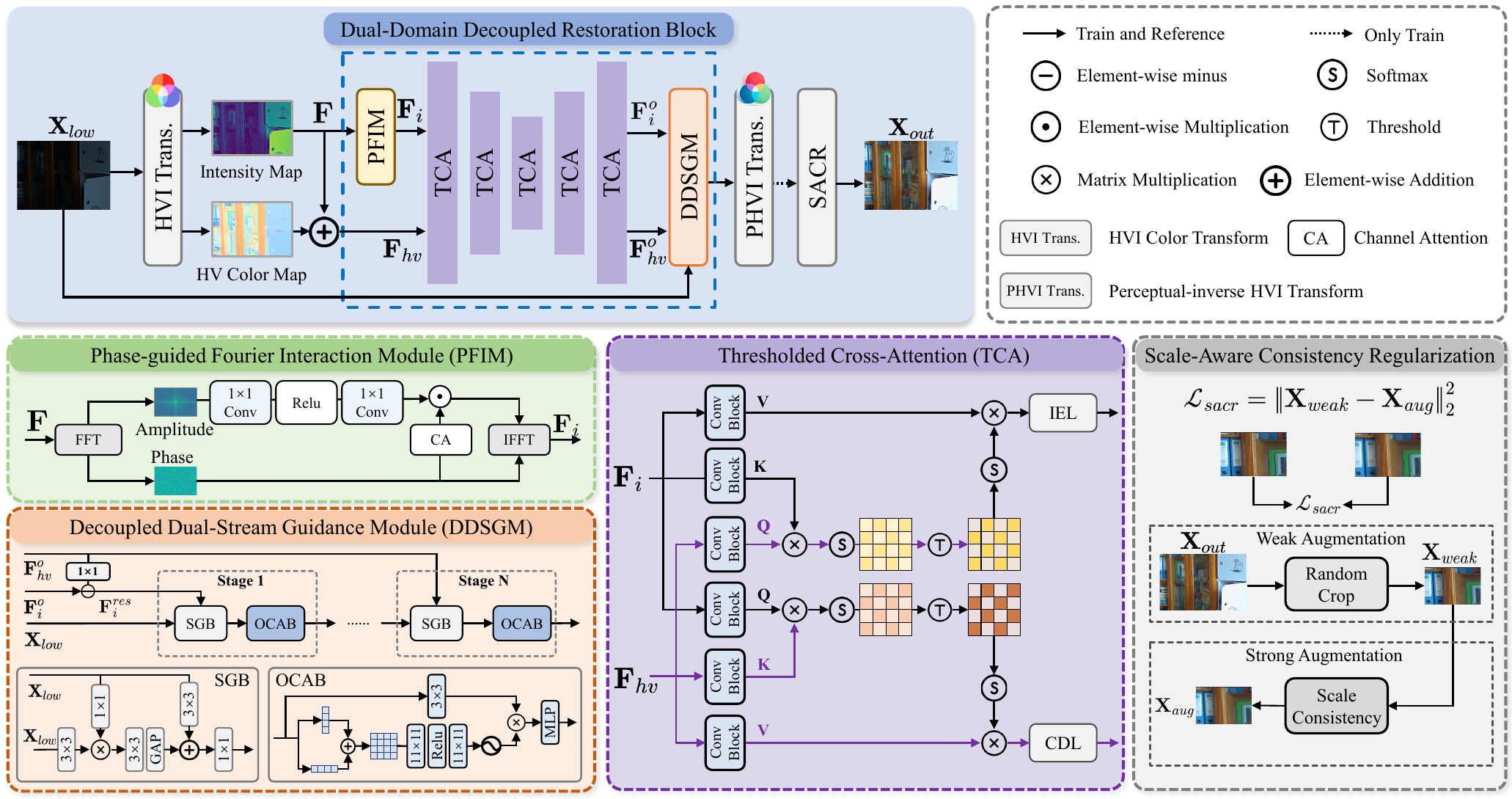}
\caption{
Overall architecture of the proposed TCA-Net. The input $\mathbf{X}_{low}$ is first mapped into the HVI space and split into an intensity stream $\mathbf{F}_{i}$ and a chromaticity stream $\mathbf{F}_{hv}$, which are processed by the Dual-Domain Decoupled Restoration Block (DDRB) and transformed back by the perceptual-inverse HVI transform to produce $\mathbf{X}_{out}$. Within the DDRB, the Phase-guided Fourier Interaction Module (PFIM) provides a structure-aware initialization for the intensity stream, a stack of Thresholded Cross-Attention (TCA) layers performs reliability-aware intensity--chromaticity fusion, and the Decoupled Dual-Stream Guidance Module (DDSGM) reconstructs the result via residual intensity guidance. The lower-left panels detail PFIM and the SGB/OCAB blocks of DDSGM; the middle panel details the TCA layer; the right panel illustrates the Scale-Aware Consistency Regularization (SACR) used only during training. Legend symbols are defined at the top right.}
\label{fig:model}
\end{figure*}

As shown in Fig.~\ref{fig:model}, given a low-light input image $\mathbf{X}_{low}$, TCA-Net first transforms it into the HVI color space to decouple the intensity component from the chromaticity components. The intensity stream is first encoded into shallow intensity features, which are enhanced by the proposed Phase-guided Fourier Interaction Module (PFIM) to provide a structure-aware brightness initialization.
Then, Thresholded Cross-Attention (TCA) is employed to perform reliability-aware interaction between the intensity and chromaticity streams by pruning low-confidence cross-stream dependencies. Finally, the Decoupled Dual-Stream Guidance Module (DDSGM) reconstructs the enhanced image by using residual intensity features to reduce chromaticity-correlated interference. These three modules are organized into a Dual-Domain Decoupled Restoration Block (DDRB), as shown in Fig.~\ref{fig:model}, after which a perceptual-inverse HVI transform maps the result back to the RGB domain to produce the output $\mathbf{X}_{out}$.

During training, the model is optimized with both external supervised reconstruction loss and internal SCAR. The total loss is formulated as:
\begin{equation}
\mathcal{L}_{total}
=
\mathcal{L}_{sup}
+
\beta(t)\mathcal{L}_{sacr},
\label{eq:total_loss}
\end{equation}
where $\mathcal{L}_{sup}$ denotes the supervised reconstruction loss, and $\mathcal{L}_{sacr}$ denotes the proposed SACR

Specifically, the supervised loss is computed in both RGB and HVI spaces:
\begin{equation}
\mathcal{L}_{sup}
=
\mathcal{L}_{rgb}(\hat{\mathbf{X}}, \mathbf{X}_{gt})
+
\lambda_{hvi}
\mathcal{L}_{hvi}
\left(
\mathcal{H}(\hat{\mathbf{X}}),
\mathcal{H}(\mathbf{X}_{gt})
\right),
\label{eq:sup_loss}
\end{equation}
where $\hat{\mathbf{X}}$ and $\mathbf{X}_{gt}$ denote the enhanced result and the ground-truth image, respectively, and $\mathcal{H}(\cdot)$ represents the HVI transformation.

For a pair of prediction and target images $(\mathbf{X},\mathbf{Y})$, the reconstruction loss is defined as:
\begin{equation}
\begin{aligned}
\mathcal{L}_{rec}(\mathbf{X},\mathbf{Y})
=&
\lambda_{1}\|\mathbf{X}-\mathbf{Y}\|_{1}
+
\lambda_{ssim}\left(1-\mathrm{SSIM}(\mathbf{X},\mathbf{Y})\right)  \\
&+
\lambda_{edge}\left\|\Delta(\mathbf{X})-\Delta(\mathbf{Y})\right\|_{2}^{2}
+
\lambda_{per}\mathcal{L}_{per}(\mathbf{X},\mathbf{Y}),
\end{aligned}
\label{eq:rec_loss}
\end{equation}
where $\Delta(\cdot)$ denotes the Laplacian edge operator, and $\mathcal{L}_{per}$ is the VGG-based perceptual loss. Accordingly, we have
$\mathcal{L}_{rgb}=\mathcal{L}_{rec}(\hat{\mathbf{X}},\mathbf{X}_{gt})$ and
$\mathcal{L}_{hvi}=\mathcal{L}_{rec}(\mathcal{H}(\hat{\mathbf{X}}),\mathcal{H}(\mathbf{X}_{gt}))$.

\subsection{Phase-guided Fourier Interaction Module (PFIM)}

After HVI transformation, the intensity component is encoded into a shallow feature $\mathbf{F}\in\mathbb{R}^{C\times H\times W}$, on which PFIM performs frequency-domain enhancement to provide a structure-aware initialization for subsequent intensity--chromaticity fusion.

Given $\mathbf{F}$, we first project it with a $1\times1$ convolution and transform it into the frequency domain:
\begin{equation}
    \mathbf{X}_{f}
    =
    \mathcal{F}\left(\psi_{pre}(\mathbf{F})\right),
    \quad
    \mathbf{M}=|\mathbf{X}_{f}|,
    \quad
    \mathbf{\Phi}=\angle \mathbf{X}_{f},
    \label{eq:pfim_fft}
\end{equation}
where $\psi_{pre}(\cdot)$ denotes the $1\times1$ pre-projection, and $\mathbf{M}$ and $\mathbf{\Phi}$ denote the amplitude and phase components, respectively.
Since amplitude mainly reflects global intensity distribution while phase preserves structural information, we modulate the amplitude under phase guidance. Specifically, the amplitude is processed by a lightweight convolutional block, while the phase is used to generate channel-wise attention:
\begin{equation}
    \mathbf{M}_{final}
    =
    \psi_{m}(\mathbf{M})
    \odot
    \sigma\left(
    \psi_{a}\left(\mathrm{GAP}(\psi_{p}(\mathbf{\Phi}))\right)
    \right),
    \label{eq:pfim_modulation}
\end{equation}
where $\psi_m(\cdot)$ denotes amplitude modulation, $\psi_p(\cdot)$ denotes phase projection, $\psi_a(\cdot)$ denotes the channel-attention network, and $\odot$ represents element-wise multiplication.

The enhanced intensity feature is then reconstructed by inverse FFT with a residual connection:
\begin{equation}
    \mathbf{F}_{i} = 
    \psi_{out}
    \left(
    \mathcal{F}^{-1}
    \left(
    \mathbf{M}_{final}\cdot e^{j\psi_{p}(\mathbf{\Phi})}
    \right)
    +
    \mathbf{F}
    \right),
    \label{eq:pfim_output}
\end{equation}
where $\psi_{out}(\cdot)$ is the output projection. In this way, PFIM enhances the intensity representation in the frequency domain while preserving phase-guided structural cues.

\subsection{Thresholded Cross-Attention (TCA)}

Following the dual-branch HVI paradigm~\cite{yan2025hvi}, our framework separately processes the intensity and chromaticity streams. Although this decomposition alleviates color entanglement, reliable cross-stream interaction is still crucial, since noise-corrupted intensity cues or chromatic artifacts may be propagated during fusion.

For the $l$-th cross-attention layer, we denote the source and target features as $\mathbf{F}_{s}^{l}$ and $\mathbf{F}_{t}^{l}$, where $s,t\in\{i,hv\}$ and $s\neq t$. Here, $i$ and $hv$ indicate the intensity and chromaticity streams, respectively. The query is generated from the source stream, while the key and value are projected from the target stream:
\begin{equation}
\mathbf{Q}^{l}=\phi_q(\mathbf{F}_{s}^{l}), \quad
\mathbf{K}^{l}=\phi_k(\mathbf{F}_{t}^{l}), \quad
\mathbf{V}^{l}=\phi_v(\mathbf{F}_{t}^{l}),
\label{eq:tca_qkv}
\end{equation}
where $\phi_q(\cdot)$, $\phi_k(\cdot)$, and $\phi_v(\cdot)$ are convolutional projections. After reshaping $\mathbf{Q}^{l}$, $\mathbf{K}^{l}$, and $\mathbf{V}^{l}$ to $\mathbb{R}^{B\times h\times d\times HW}$, the dense channel-wise attention is computed as:
\begin{equation}
\mathbf{A}^{l}
=
\operatorname{Softmax}
\left(
\gamma^{l}\mathbf{Q}^{l}(\mathbf{K}^{l})^{\top}
\right),
\quad
\mathbf{Y}^{l}=\mathbf{A}^{l}\mathbf{V}^{l},
\label{eq:channel_attn}
\end{equation}
where $\gamma^{l}$ is a learnable temperature parameter. Since Softmax assigns non-zero weights to all target channels, weak or noise-correlated responses may still be propagated across streams.

A common sparsification strategy is Top-$\mathrm{K}$ selection, which keeps the $\mathrm{K}$ largest attention responses for each source channel:
\begin{equation}
\mathbf{M}^{\mathrm{K},l}
=
\operatorname{TopKMask}(\mathbf{A}^{l},\mathrm{K}),
\quad
\mathrm{K}=\lceil \rho d\rceil ,
\label{eq:channel_topk_mask}
\end{equation}
where $\rho$ is a predefined retention ratio. The corresponding sparse attention is:
\begin{equation}
\hat{\mathbf{A}}^{\mathrm{K},l}
=
\operatorname{Norm}
\left(
\mathbf{A}^{l}\odot\mathbf{M}^{\mathrm{K},l}
\right),
\quad
\mathbf{Y}^{\mathrm{K},l}
=
\hat{\mathbf{A}}^{\mathrm{K},l}\mathbf{V}^{l},
\label{eq:channel_topk_attn}
\end{equation}
where $\operatorname{Norm}(\cdot)$ denotes re-normalization along the target-channel dimension. However, Top-$\mathrm{K}$ imposes a fixed retention quota on all layers and source channels. This rigid selection rule may discard useful dependencies in information-dense layers while retaining weak or noise-correlated responses in diffuse attention distributions.

To address this limitation, we propose Thresholded Cross-Attention (TCA), which performs confidence-based pruning on the normalized attention map. Instead of retaining a fixed number of responses, TCA preserves interactions whose attention confidence exceeds a fixed threshold $\tau$:
\begin{equation}
\mathbf{M}^{\tau,l}
=
\operatorname{Mask}
\left(
\mathbf{A}^{l}\geq \tau
\right)
\vee
\operatorname{OneHotArgmax}
\left(
\mathbf{A}^{l}
\right),
\label{eq:channel_tca_mask}
\end{equation}
here, $\operatorname{OneHotArgmax}(\cdot)$ is computed along the target-channel dimension and keeps the largest response for each source channel. This fallback prevents an all-zero mask when all responses are below $\tau$. The pruned attention is then computed as:
\begin{equation}
\hat{\mathbf{A}}^{\tau,l}
=
\operatorname{Norm}
\left(
\mathbf{A}^{l}\odot\mathbf{M}^{\tau,l}
\right),
\quad
\mathbf{Y}^{\tau,l}
=
\hat{\mathbf{A}}^{\tau,l}\mathbf{V}^{l}.
\label{eq:channel_tca_attn}
\end{equation}
where $\operatorname{Norm}(\cdot)$ re-normalizes the masked attention along the target-channel dimension with a small numerical constant in the denominator.

In our implementation, we set $\tau=0.025$ according to the ablation analysis. The threshold is applied to Softmax-normalized attention probabilities rather than raw feature activations, so it operates on a comparable confidence scale across inputs and layers. Although $\tau$ is shared across layers, the retained cardinality and sparse pattern remain input- and layer-adaptive because they depend on the attention distribution of each input and layer. 
As shown in Fig.~\ref{fig:attn_hist}, the attention distributions vary across layers and datasets, while TCA consistently preserves high-confidence responses and suppresses low-confidence ones according to the same probability-level confidence criterion.
The refined features are then fed into the Illumination Estimation Layer (IEL) and Chromaticity Denoising Layer (CDL)~\cite{yan2025hvi} to generate the enhanced intensity and chromaticity components.

\subsection{Decoupled Dual-Stream Guidance Module (DDSGM)}

After TCA, the refined intensity and chromaticity features still need to be effectively fused for final reconstruction. Direct concatenation may mix illumination and color information again, leading to chromatic interference. To alleviate this issue, we propose the Decoupled Dual-Stream Guidance Module (DDSGM), which explicitly constructs a residual intensity feature to reduce chromaticity-related leakage.

Given the refined intensity feature $\mathbf{F}_{i}^{o}$ and chromaticity feature $\mathbf{F}_{hv}^{o}$, we first generate a learnable adaptive map $\mathbf{M}_{\mu}$ from the chromaticity stream and use it to remove chromaticity-correlated components from the intensity feature:
\begin{gather}
    \mathbf{M}_{\mu} = \sigma(\mathrm{Conv}(\mathbf{F}_{hv}^{o})), \label{eq:m_definition} \\
    \mathbf{F}_{i}^{res} = \mathbf{F}_{i}^{o} - \mathbf{M}_{\mu} \odot \mathbf{F}_{hv}^{o}, \label{eq:f_res}
\end{gather}
where $\sigma(\cdot)$ denotes the Sigmoid function and $\odot$ represents element-wise multiplication. The residual intensity feature $\mathbf{F}_{i}^{res}$ provides cleaner illumination guidance by suppressing chromaticity-correlated interference.

DDSGM then adopts a cascaded two-stage reconstruction process. The first stage uses $\mathbf{F}_{i}^{res}$ to guide illumination restoration, while the second stage uses $\mathbf{F}_{hv}^{o}$ to refine chromaticity details. To make the data flow explicit, we denote the reconstruction feature before stage $k$ as $\mathbf{U}_{k-1}$, with $\mathbf{U}_{0}=\mathrm{Conv}_{3\times3}(\mathbf{X}_{low})$, and define the stage guidance as $\mathbf{G}_{1}=\mathbf{F}_{i}^{res}$ and $\mathbf{G}_{2}=\mathbf{F}_{hv}^{o}$. Each stage consists of a Spatial Guidance Block (SGB) for feature modulation and an Oriented Context Aggregation Block (OCAB) for spatial context modeling. For the $k$-th stage, SGB performs guidance-aware modulation as:
\begin{align}
\mathbf{S}_{k} &= \mathrm{Conv}_{3\times3}(\mathbf{U}_{k-1}) 
\odot \mathrm{Conv}_{1\times1}(\mathbf{G}_{k}), \label{eq:sgb_s}\\
\mathbf{U}'_{k} &= \mathrm{Conv}_{3\times3}(\mathbf{U}_{k-1}) 
\oplus \mathrm{GAP}(\mathrm{Conv}_{3\times3}(\mathbf{S}_{k})).
\label{eq:sgb_s_prime}
\end{align}

After modulation, OCAB aggregates horizontal and vertical contextual information using dual-path pooling and large-kernel strip convolutions~\cite{ni2024context}, producing $\mathbf{U}_{k}=\operatorname{OCAB}(\mathbf{U}'_{k})$. 
The second-stage output is finally projected to the enhanced HVI representation. This formulation separates the illumination-guided and chromaticity-guided reconstruction paths while keeping their inputs and outputs well defined.

\subsection{Scale-Aware Consistency Regularization (SACR)} 

To improve the structural robustness of the restored image under scale perturbations, we introduce a self-supervised Scale-Aware Consistency Regularization (SACR) during training. SACR is applied at the output level: it encourages the enhanced result to remain stable under mild scale degradation, while the supervised reconstruction loss still provides the main constraint for recovering fine-grained details.

Given the restored image $\hat{\mathbf{X}}$, we first generate a weakly augmented output view and its scale-perturbed counterpart:
\begin{equation}
\mathbf{X}_{weak} = \mathrm{T}_{weak}(\hat{\mathbf{X}}), \quad
\mathbf{X}_{aug} = \mathrm{T}_{aug}(\mathbf{X}_{weak}),
\label{eq:sacr_aug}
\end{equation}
where $\mathrm{T}_{weak}$ denotes weak augmentation implemented by random cropping, and $\mathrm{T}_{aug}$ denotes scale perturbation implemented by downsampling followed by bilinear upsampling to the original crop size. In our implementation, the weak augmentation crops a $128\times128$ patch, and the scale perturbation adopts $0.5\times$ downsampling followed by upsampling.

The SACR loss is defined as the $\ell_2$ distance between the weakly augmented output and its scale-perturbed version:
\begin{equation}
\mathcal{L}_{sacr}
=
\left\|
\mathbf{X}_{weak}
-
\mathbf{X}_{aug}
\right\|_{2}^{2}.
\label{eq:sacr_loss}
\end{equation}

As defined in Eq.~\eqref{eq:total_loss}, SACR is used as an auxiliary regularization term together with the supervised reconstruction loss. Its weight follows a cosine decay schedule:
\begin{equation}
\beta(t)
=
\beta_{0}
\cdot
\frac{1+\cos(\pi t/T)}{2},
\label{eq:beta_decay}
\end{equation}
where $\beta_{0}$ is the initial regularization weight, $t$ denotes the current training iteration, and $T$ is the total number of training iterations. 
This schedule gradually relaxes the output-level scale-consistency constraint, reducing the risk of over-smoothing while helping the model maintain structural stability.

\begin{table*}[!t]
\centering
\caption{Quantitative results in terms of PSNR$\uparrow$, SSIM$\uparrow$, and LPIPS$\downarrow$ on the LOL-v1 and LOL-v2 datasets. The best and second-best results are highlighted in \textcolor{red}{red} and \textcolor{blue}{blue}, respectively.}
\label{tab:performance_comparison}
{
\begin{tabular}{@{}l|c|c c c|c c c|c c c@{}}
\toprule
\multirow{2}{*}{\textbf{Methods}} 
& \multirow{2}{*}{\textbf{Color Model}} 
& \multicolumn{3}{c}{\textbf{LOL-v1}} 
& \multicolumn{3}{|c}{\textbf{LOL-v2-Real}} 
& \multicolumn{3}{|c}{\textbf{LOL-v2-Synthetic}} \\
\cmidrule(lr){3-5} 
\cmidrule(lr){6-8} 
\cmidrule(lr){9-11}
& & PSNR$\uparrow$ & SSIM$\uparrow$ & LPIPS$\downarrow$ 
& PSNR$\uparrow$ & SSIM$\uparrow$ & LPIPS$\downarrow$ 
& PSNR$\uparrow$ & SSIM$\uparrow$ & LPIPS$\downarrow$ \\
\midrule

RetinexNet \cite{wei2018deep}  & Retinex & 18.915 & 0.427 & 0.470 & 16.097 & 0.401 & 0.543 & 17.137 & 0.762 & 0.255 \\
KinD \cite{zhang2019kindling}  & Retinex & 23.018 & 0.843 & 0.156 & 17.544 & 0.669 & 0.375 & 18.320 & 0.796 & 0.252 \\
ZeroDCE \cite{guo2020zero}  & RGB & 21.880 & 0.640 & 0.335 & 16.059 & 0.580 & 0.313 & 17.712 & 0.815 & 0.169 \\
RUAS \cite{liu2021retinex} & Retinex & 18.654 & 0.518 & 0.270 & 15.326 & 0.488 & 0.310 & 13.765 & 0.638 & 0.305 \\
LLFlow \cite{wang2022low} & RGB & 24.060 & \textcolor{blue}{0.860} & 0.136 & 17.433 & 0.831 & 0.176 & 24.807 & 0.919 & 0.067 \\
EnGAN \cite{jiang2021enlightengan} & RGB & 20.003 & 0.691 & 0.317 & 18.230 & 0.617 & 0.309 & 16.570 & 0.734 & 0.220 \\
SNR-Aware \cite{xu2022snr} & RGB & 24.610 & 0.842 & 0.152 & 21.480 & 0.849 & 0.163 & 24.140 & 0.928 & 0.056 \\
Bread \cite{guo2023low} & YCbCr & 22.920 & 0.836 & 0.155 & 20.830 & 0.847 & 0.174 & 17.630 & 0.919 & 0.091 \\
PairLIE \cite{fu2023learning} & Retinex & 23.526 & 0.755 & 0.248 & 19.885 & 0.778 & 0.317 & 19.074 & 0.794 & 0.230 \\
Restormer~\cite{zamir2022restormer} & RGB & 22.427 & 0.823 & 0.147&18.600 &0.789 &-- &21.410 &0.831 & --\\
LLFormer \cite{wang2023ultra} & RGB & 23.649 & 0.816 & 0.167 & 20.056 & 0.792 & 0.211 & 24.038 & 0.909 & 0.066 \\
RetinexFormer \cite{cai2023retinexformer} & Retinex & \textcolor{blue}{25.161} & 0.845 & 0.129 & 22.794 & 0.840 & 0.171 & 25.670 & 0.930 & 0.059 \\
QuadPrior \cite{wang2024zero} & Kubelka-Munk & 22.849 & 0.800 & 0.201 & 20.592 & 0.811 & 0.202 & 16.108 & 0.758 & 0.114 \\
END \cite{wang2024extracting} &YCbCr &24.740 & 0.837 & 0.133 & - & - & - & - & - & -  \\
CIDNet \cite{yan2025hvi} & HVI & 23.809 & 0.857 & \textcolor{blue}{0.079} & \textcolor{blue}{23.900} & \textcolor{blue}{0.865} & \textcolor{red}{0.108} & \textcolor{blue}{25.705} & \textcolor{blue}{0.942} & \textcolor{blue}{0.045} \\

\rowcolor[HTML]{EDEDED}
\bf Ours & HVI & \textcolor{red}{25.322} & \textcolor{red}{0.876} & \textcolor{red}{0.069} & \textcolor{red}{24.171} & \textcolor{red}{0.872} & \textcolor{blue}{0.109} & \textcolor{red}{26.023} & \textcolor{red}{0.944} & \textcolor{red}{0.040} \\
\bottomrule
\end{tabular}}
\end{table*}

\section{Experiments}
\subsection{Datasets}\label{sec:exp_datasets}
In our experiments, we evaluate the proposed method on five widely used LLIE benchmarks: LOL-v1~\cite{wei2018deep}, LOL-v2-Real~\cite{yang2021sparse}, LOL-v2-Synthetic~\cite{yang2021sparse}, LSRW-Huawei~\cite{hai2023r2rnet}, and Sony-Total-Dark~\cite{chen2018learning}. The details of these datasets and training settings are summarized as follows.

\textbf{LOL-v1} is a real-world paired benchmark for LLIE, consisting of 500 low-/normal-light image pairs, including 485 training pairs and 15 testing pairs. During training, we crop the images into $256\times256$ patches and train the model for 2000 epochs with a batch size of 4.

\textbf{LOL-v2-Real} extends LOL-v1 with more diverse scenes and challenging illumination conditions. It contains 789 real-world image pairs, including 689 training pairs and 100 testing pairs. We crop the training images into $400\times400$ patches and train for 1500 epochs with a batch size of 6.

\textbf{LOL-v2-Synthetic} provides large-scale synthetic paired data with accurate ground truth. It consists of 1,000 image pairs, with 900 pairs for training and 100 pairs for testing. We use $256\times256$ training patches and train for 1500 epochs with a batch size of 4.

\textbf{LSRW-Huawei} is a real-world low-light benchmark collected for mobile photography scenarios. It contains 2,450 training pairs and 30 testing pairs captured by Huawei mobile devices under diverse natural lighting conditions.

\textbf{Sony-Total-Dark} is used to evaluate the robustness of the model under extremely low-light conditions. It is built from a modified subset of SID, where raw images are converted to sRGB without gamma correction to simulate severe underexposure. We crop the training images into $256\times256$ patches and train for 1000 epochs with a batch size of 4.

\subsection{Implementation Details}

The proposed model is implemented in PyTorch and optimized using the Adam optimizer with $\beta_1=0.9$ and $\beta_2=0.999$. The initial learning rate is set to $1\times10^{-4}$ and decayed to a minimum of $1\times10^{-7}$ following a cosine annealing schedule over the whole training process. To stabilize training, gradient clipping is applied with a maximum $\ell_2$ norm of $0.01$. All models are trained with a batch size of $4$ on $256\times256$ patches, except where dataset-specific settings are noted in Sec.~\ref{sec:exp_datasets}. 
The supervised reconstruction loss in Eq.~\eqref{eq:rec_loss} combines an $\ell_1$ term, an SSIM term, an edge term, and a VGG-based perceptual term, with weights $\lambda_1=1.0$, $\lambda_{ssim}=0.5$, $\lambda_{edge}=50.0$, and $\lambda_{per}=1\times10^{-2}$, respectively. This reconstruction loss is applied in both the RGB and HVI spaces, where the HVI term is weighted by $\lambda_{hvi}=1.0$ as in Eq.~\eqref{eq:sup_loss}. The scale-aware consistency regularization is weighted by $\beta(t)$ following the cosine decay schedule in Eq.~\eqref{eq:beta_decay}. All experiments are conducted on a single NVIDIA GeForce RTX 4090 GPU.

\subsection{Evaluation Metrics}

We evaluate enhancement performance using fidelity, perceptual, and color-related metrics, including PSNR, SSIM~\cite{SSIM}, LPIPS~\cite{LPIPS}, and chromaticity error, which measure pixel-level accuracy, structural consistency, perceptual similarity, and color restoration quality.

In addition to general image fidelity, we evaluate color restoration accuracy using a chromaticity-only error derived from the CIE Lab color-difference formulation~\cite{robertson1977cie}. The enhanced image $\mathbf{X}_{out}$ and the ground-truth image $\mathbf{X}_{gt}$ are first converted from RGB to Lab space. Since the $L^{*}$ channel mainly represents luminance, we compute the color discrepancy only on the chromaticity channels $a^{*}$ and $b^{*}$:
\begin{equation}
    E_{ab}(p)
    =
    \sqrt{
    \left(
    a_{out}(p)-a_{gt}(p)
    \right)^2
    +
    \left(
    b_{out}(p)-b_{gt}(p)
    \right)^2
    },
    \label{eq:eab}
\end{equation}
where $p$ denotes a spatial pixel location. $a_{out}(p)$ and $b_{out}(p)$ are the chromaticity values of the enhanced image, while $a_{gt}(p)$ and $b_{gt}(p)$ are those of the ground truth. Different from the full CIE $\Delta E_{ab}^{*}$ color difference, this metric excludes the luminance channel and focuses only on chromaticity discrepancy. A lower $E_{ab}$ value indicates more faithful chromaticity restoration. In our experiments, we first compute the average $E_{ab}$ over all pixels of each test image, and then report the mean and standard deviation over the whole test set to measure both overall chromatic accuracy and color stability.

\subsection{Main Results}
\textbf{Results on LOL datasets.} We conduct experiments on the widely-used LOL-v1 and LOL-v2 datasets to evaluate the performance of our proposed method. For a comprehensive comparison, we evaluate our approach against fifteen representative and recent low-light image enhancement methods, including RetinexNet \cite{wei2018deep}, KinD \cite{zhang2019kindling}, ZeroDCE \cite{guo2020zero}, RUAS \cite{liu2021retinex}, LLFlow \cite{wang2022low}, EnGAN \cite{jiang2021enlightengan}, SNR-Aware \cite{xu2022snr}, Bread \cite{guo2023low}, PairLIE \cite{fu2023learning}, Restormer \cite{zamir2022restormer}, LLFormer \cite{wang2023ultra}, RetinexFormer \cite{cai2023retinexformer}, QuadPrior \cite{wang2024zero}, END \cite{wang2024extracting}, and CIDNet \cite{yan2025hvi}. 

\begin{figure*}[!t]
    \centering
	\includegraphics[width=\textwidth]{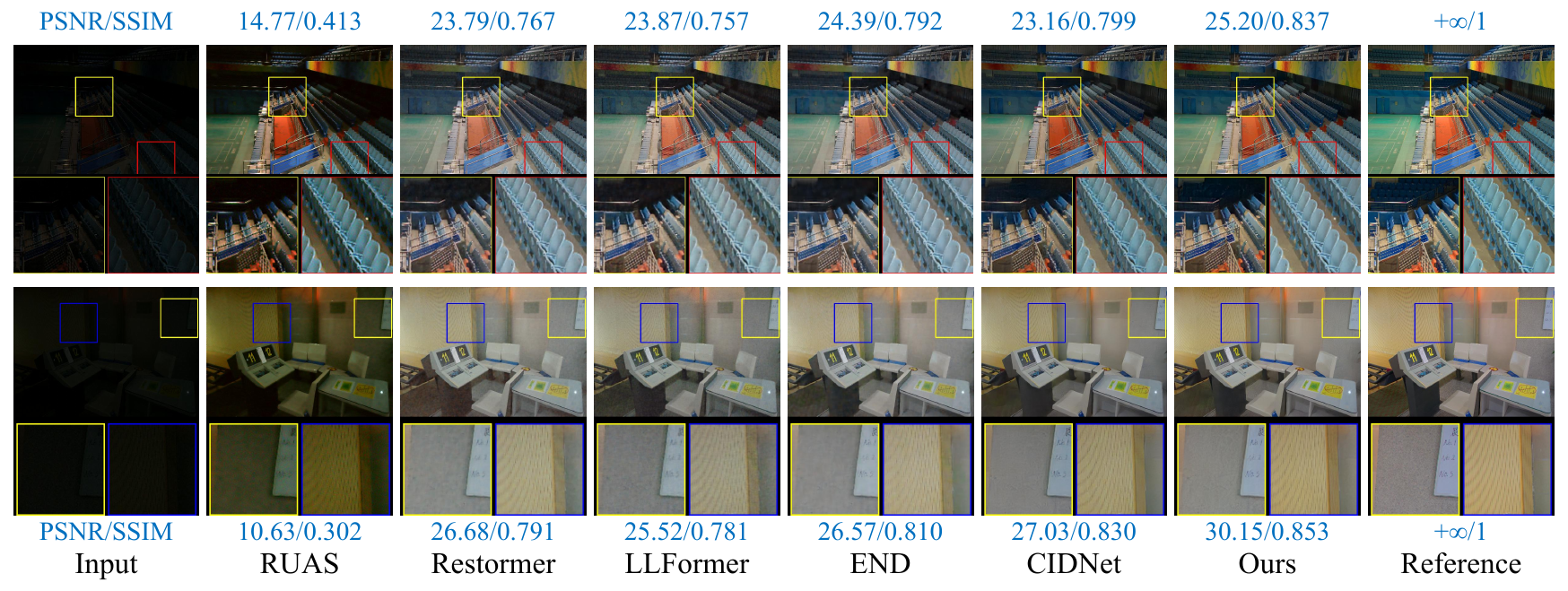}
    \caption{Visual comparison of the enhanced images yielded by different methods on LOL-v1.}
    \label{fig:visuallolv1}
\end{figure*}

\begin{figure*}[!t]
	\centering
	\includegraphics[width=\textwidth]{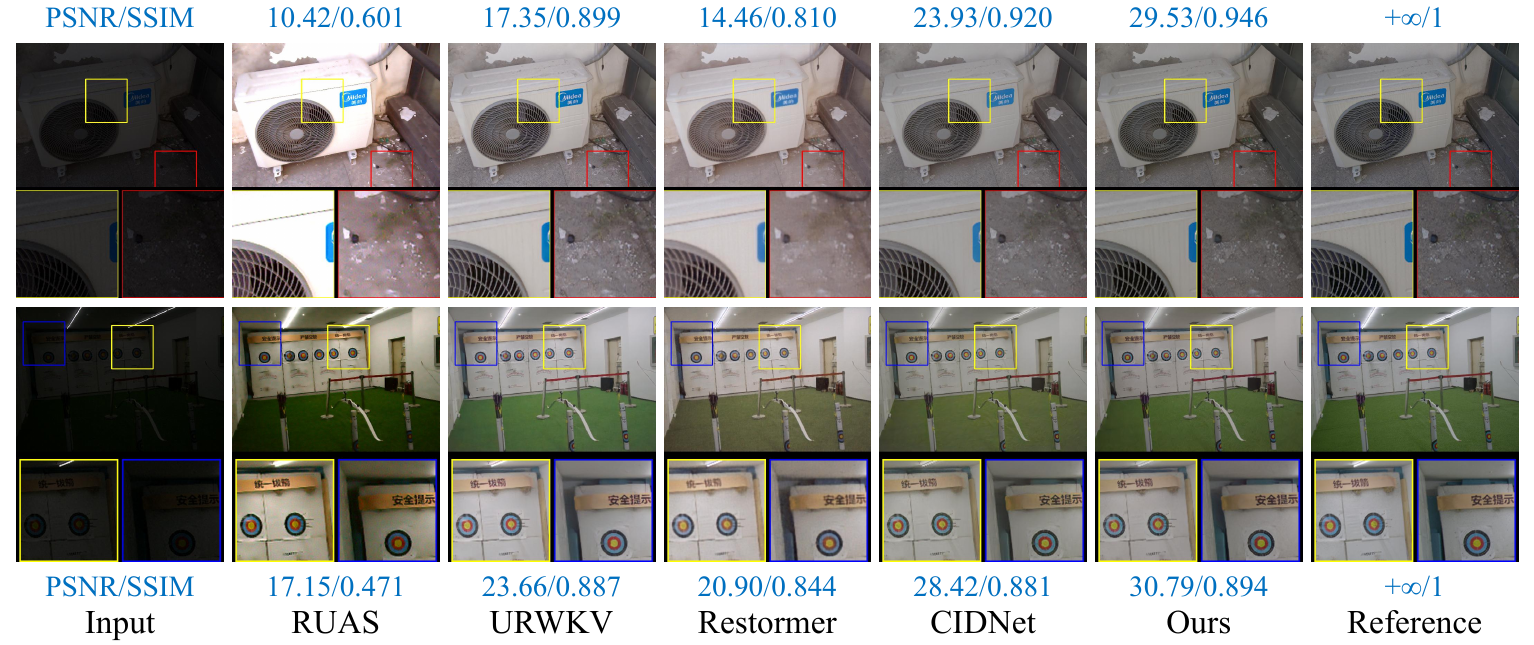}
	\caption{Visual comparison of the enhanced images yielded by different methods on LOL-v2-Real.}
	\label{fig:visuallolv2real}
\end{figure*}

As summarized in Table~\ref{tab:performance_comparison}, our method ranks first or remains competitive across most metrics, with only a marginal LPIPS difference on LOL-v2-Real compared with CIDNet. Furthermore, qualitative comparisons are illustrated in Fig.~\ref{fig:visuallolv1} (LOL-v1) and Fig.~\ref{fig:visuallolv2real} (LOL-v2-Real). Compared with other existing methods, our approach restores more natural colors and preserves finer structural details.

\begin{figure*}[!htbp]
    \centering
    \includegraphics[width=\linewidth]{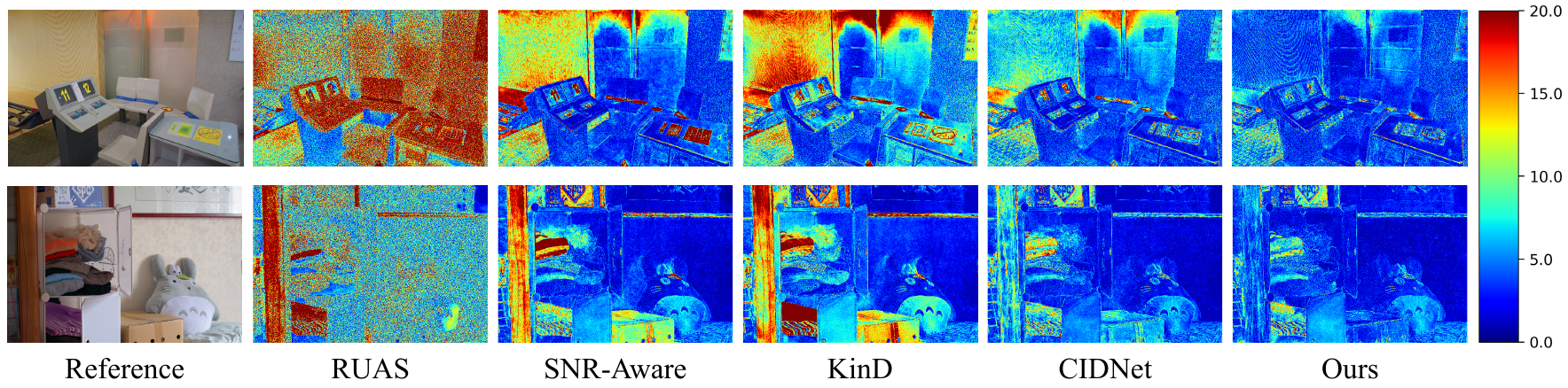}
\caption{Chromaticity error visualization on representative LOL-v1 samples. Higher responses indicate larger chromaticity deviations between the enhanced result and the ground truth. Compared with other methods, our method produces fewer high-error regions, suggesting more faithful chromaticity restoration.}
    \label{fig:chrom_error_lolv1}
\end{figure*}

\begin{table}[!htbp]
\centering
\caption{$E_{ab}$ chromaticity error comparison on the LOL-v1 test set.}
\label{tab:eab_error}
\begin{tabular}{lcc}
\toprule
\rowcolor[HTML]{EDEDED}
Model & Avg. $E_{ab}\downarrow$ & Std. $E_{ab}\downarrow$ \\
\midrule
RUAS  \cite{liu2021retinex}    & 12.132 & 2.723 \\
SNR-Aware \cite{xu2022snr}  & 6.797  & 3.027\\
KinD \cite{zhang2019kindling} & 5.618  & 3.278 \\
CIDNet \cite{yan2025hvi} & 5.459  & 2.431 \\

\rowcolor[HTML]{EDEDED}
\textbf{Ours}  & \textbf{5.058}  & \textbf{2.105} \\
\bottomrule
\end{tabular}
\end{table}

\emph{Color fidelity analysis on LOL-v1.} To further evaluate chromaticity restoration, we compute the chromaticity error $E_{ab}$ on the LOL-v1 test set following Eq.~\eqref{eq:eab}. Since $E_{ab}$ measures the discrepancy only on the chromaticity channels, it provides a direct evaluation of color restoration accuracy while excluding luminance-induced variations. As shown in Table~\ref{tab:eab_error}, our method achieves the lowest average color error and the smallest standard deviation among all compared methods. Compared with CIDNet, which also benefits from the HVI color space, our method further reduces both Avg. $E_{ab}$ and Std. $E_{ab}$, demonstrating more accurate and stable chromaticity restoration. We further visualize the chromaticity error maps in Fig.~\ref{fig:chrom_error_lolv1}. RUAS introduces severe color deviations over large image regions, while SNR-Aware and KinD still exhibit noticeable chromatic errors in textured and low-light areas. CIDNet effectively reduces the overall color discrepancy, but local color inconsistency can still be observed around complex structures. In contrast, our method produces broader low-error regions and fewer high-error responses, especially around dark areas, texture boundaries, and chromatically complex regions. These results further confirm the advantage of the proposed method in faithful color recovery.

\textbf{Results on Sony-Total-Dark dataset.}
To evaluate the robustness of our framework in extremely low-light environments, we conduct experiments on the Sony subset of the SID dataset. This dataset is particularly challenging due to near-total darkness and severe noise degradation. We compare our method with several representative LLIE methods, including RetinexNet~\cite{wei2018deep}, ZeroDCE~\cite{guo2020zero}, RUAS~\cite{liu2021retinex}, URetinexNet~\cite{Wu_2022_CVPR}, LLFlow~\cite{wang2022low}, and CIDNet~\cite{yan2025hvi}.

\begin{figure*}[!t]
    \centering
    \includegraphics[width=\linewidth]{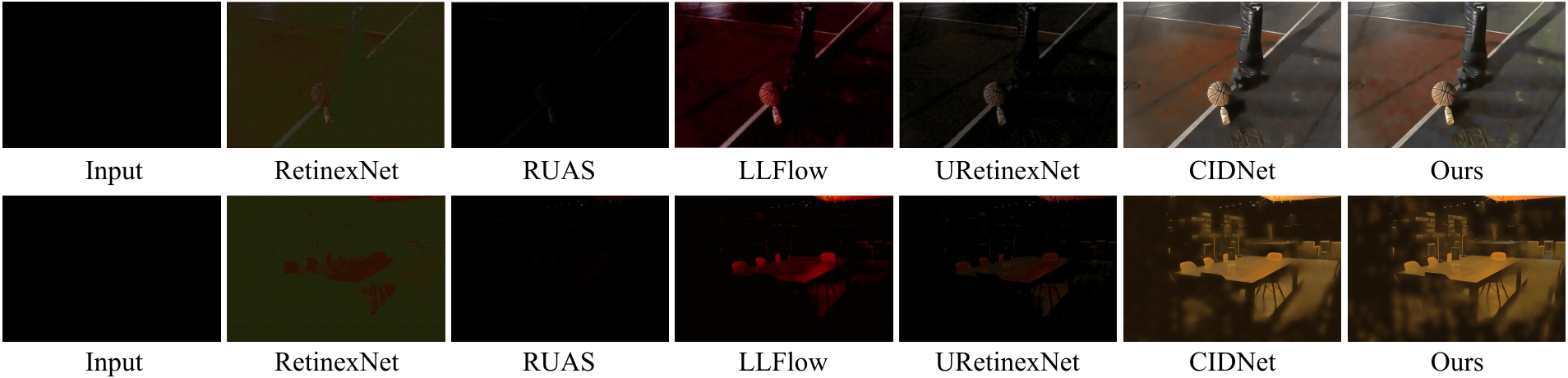}
    \caption{Visual comparison of the enhanced images yielded by different methods on Sony-Total-Dark.}
    \label{fig:SID}
\end{figure*}

As shown in Table~\ref{tab:comparison_sid}, our method obtains the highest reported PSNR of 23.03 dB and SSIM of 0.677 among the compared methods. Compared with CIDNet, our method improves the PSNR by 0.13 dB while slightly improving SSIM from 0.676 to 0.677. These results indicate that the proposed method achieves stronger pixel-level reconstruction fidelity while maintaining competitive structural similarity under extremely low-light conditions. Qualitative comparisons are shown in Fig.~\ref{fig:SID}.

\begin{table}[!htbp]
\centering
\caption{Quantitative results on the SID dataset. The top-ranking score is shown in \textcolor{red}{red}, and the second-best is shown in \textcolor{blue}{blue}.}
\label{tab:comparison_sid}
\begin{tabular}{lcc}

\toprule
\rowcolor[HTML]{EDEDED}
Methods  & PSNR$\uparrow$ & SSIM$\uparrow$ \\

\midrule
RetinexNet \cite{wei2018deep} & 15.70 & 0.395 \\
ZeroDCE \cite{guo2020zero}  & 14.09 & 0.090 \\
RUAS \cite{liu2021retinex} & 12.62 & 0.081 \\
URetinexNet \cite{Wu_2022_CVPR} & 15.52 & 0.323 \\
LLFlow \cite{wang2022low}  & 16.23 & 0.367 \\
CIDNet \cite{yan2025hvi} & \textcolor{blue}{22.90} &\textcolor{blue}{0.676} \\

\rowcolor[HTML]{EDEDED}
\textbf{Ours} & \textcolor{red}{23.03} & \textcolor{red}{0.677} \\ 

\bottomrule
\end{tabular}
\end{table}

\textbf{Results on LSRW-Huawei dataset.} 
To further assess the generalization performance of our method, we conduct experiments on the LSRW-Huawei dataset, which is widely recognized for its challenging real-world low-light scenarios. We compare our approach with a range of representative methods, including KinD \cite{zhang2019kindling}, RUAS \cite{liu2021retinex}, EnGAN~\cite{jiang2021enlightengan}, SNR-Aware \cite{xu2022snr}, RetinexFormer \cite{cai2023retinexformer}, FourLLIE \cite{wang2023fourllie}, UHDFour \cite{li2023embedding}, Bread \cite{guo2023low}, and DarkIR \cite{Feijoo_2025_CVPR}.

\begin{figure*}[!t]
    \centering
    \includegraphics[width=\linewidth]{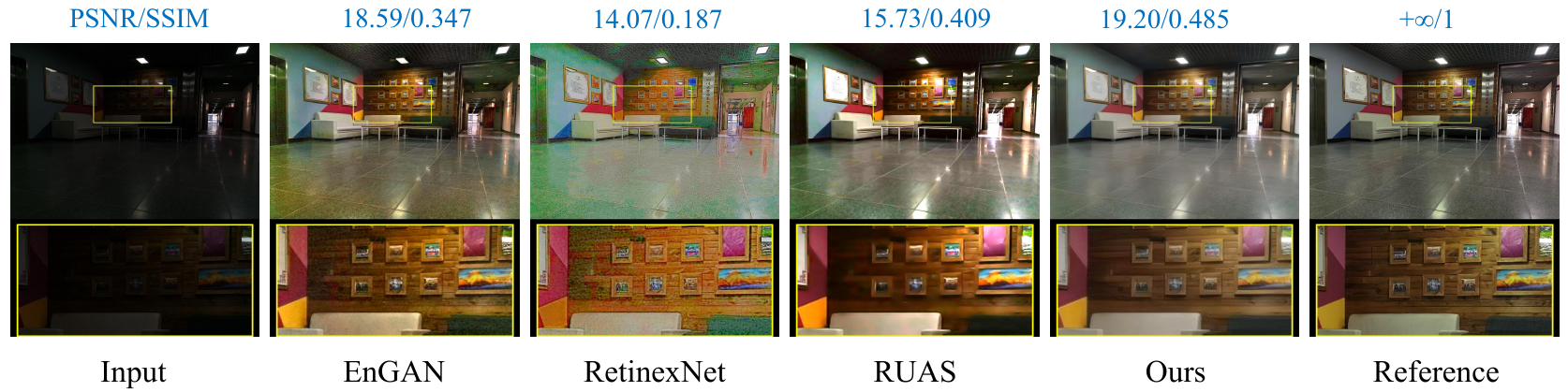}
    \caption{Visual comparison of the enhanced images yielded by different methods on LSRW-Huawei.}
    \label{fig:huawei}
\end{figure*}

As summarized in Table~\ref{tab:lsrw_huawei}, our method obtains the highest PSNR in this comparison and maintains comparable SSIM to the leading methods, demonstrating its effectiveness in enhancing image brightness while preserving structural consistency. This further confirms the adaptability of our approach to complex real-world low-light conditions. Furthermore, we visualize several representative methods in Fig.~\ref{fig:huawei}.

\begin{table}[!htbp]
\centering
\caption{Quantitative results on the LSRW-Huawei dataset. The top-ranking score is shown in \textcolor{red}{red}, and the second-best is shown in \textcolor{blue}{blue}.}
\label{tab:lsrw_huawei}
\begin{tabular}{l c c c}

\toprule
\rowcolor[HTML]{EDEDED}
Methods & Venue & PSNR$\uparrow$ & SSIM$\uparrow$ \\

\midrule
KinD \cite{zhang2019kindling} & MM'19 & 16.58 & 0.569 \\
EnGAN~\cite{jiang2021enlightengan} & TIP'21 & 16.31 & 0.470 \\
RUAS  \cite{liu2021retinex} & CVPR'21 &14.44 & 0.428 \\
SNR-Aware \cite{xu2022snr} & CVPR'22 & 20.67 & 0.591 \\
RetinexFormer \cite{cai2023retinexformer} & ICCV'23 & \textcolor{blue}{21.23} & \textcolor{red}{0.631} \\
FourLLIE \cite{wang2023fourllie} & MM'23 & 21.11 & 0.626 \\
UHDFour \cite{li2023embedding} & ICLR'23 & 19.39 & 0.600\\
Bread \cite{guo2023low} &IJCV'23 &19.20&0.618\\
DarkIR \cite{Feijoo_2025_CVPR} & CVPR'25 & 18.93 & 0.583 \\

\rowcolor[HTML]{EDEDED}
\textbf{Ours} & -- & \textcolor{red}{21.35} & \textcolor{blue}{0.627} \\

\bottomrule
\end{tabular}
\end{table}

\subsection{Task-Relevant Visual Analysis}

Beyond standard restoration metrics, low-light enhancement should also preserve visual cues that are relevant to high-level visual understanding, such as stable object boundaries, sufficient foreground--background contrast, and reliable local textures. An enhancement result that only increases brightness may still be suboptimal if it introduces color shifts, over-smoothed structures, or noise-amplified regions.

\begin{figure*}[!t]
    \centering
    \includegraphics[width=\linewidth]{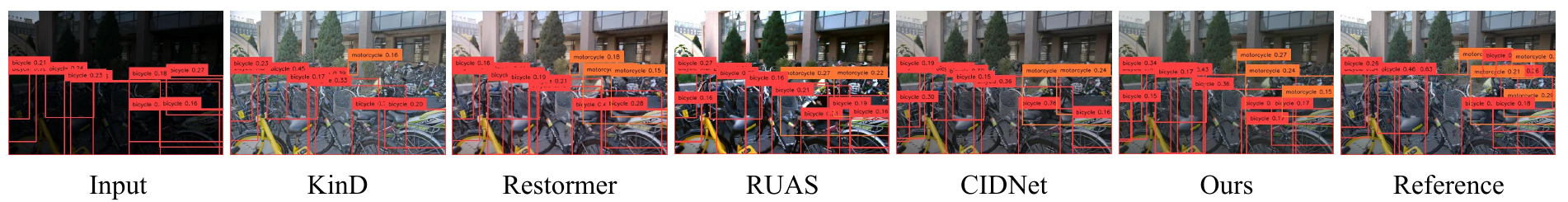}
    \caption{Task-relevant visual comparison on a representative LOL-v2-Real image using YOLOv10 detection model. The text above each box denotes the detected category and the number denotes the confidence score. Compared with existing methods, our method restores clearer object structures, more stable colors, and better foreground--background separation.}
    \label{fig:task_relevant_real715}
\end{figure*}

As shown in Fig.~\ref{fig:task_relevant_real715}, the low-light input contains weak structural responses and poor local contrast, making scene contents difficult to distinguish for YOLOv10~\cite{wang2024yolov10}. KiND and Restormer improve the overall brightness, but some object boundaries remain unclear. RUAS produces stronger contrast, yet its aggressive enhancement may amplify edges and noise-like textures. CIDNet gives a relatively stable visual appearance, but some small structures are still weakened by residual darkness and over-smoothed details. In contrast, our method recovers more coherent illumination while maintaining sharper local structures and more natural chromaticity, achieving better detection of bicycles and motorcycles. The enhanced results suggest that the proposed intensity-chromaticity fusion helps preserve task-relevant semantic and structural cues in addition to improving pixel-level reconstruction.

\subsection{Computational Complexity and Inference Efficiency}

To further assess the practical efficiency of the proposed method, we compare the model complexity and inference time with representative LLIE methods, as reported in Table~\ref{tab:complexity_runtime}. Specifically, we report the number of parameters, FLOPs, and single-image inference time to provide a comprehensive view of computational cost. All measurements are conducted with an input resolution of $256 \times 256$, and the inference time is evaluated on the same NVIDIA GeForce RTX 4090 platform used in our experiments.

\begin{table}[H]
\centering
\caption{Comparison of model complexity and inference efficiency. The inference time is reported in milliseconds.}
\label{tab:complexity_runtime}
{
\begin{tabular}{lccc}
\toprule
\rowcolor[HTML]{EDEDED}
{Method} & {Params (M)} & {FLOPs (G)} & {Time (ms)} \\
\midrule
RetinexFormer~\cite{cai2023retinexformer} & 1.53  & 15.85  & 8.96  \\
LLFlow~\cite{wang2022low}                 & 37.06 & 533.34 & 52.40  \\
LLFormer~\cite{wang2023ultra}             & 24.55 & 44.07  & 30.37  \\
CIDNet~\cite{yan2025hvi}                  & 1.88  & 7.57   & 11.20 \\
EnGAN~\cite{jiang2021enlightengan} & 8.64  & 32.85  & 1.23  \\

\rowcolor[HTML]{EDEDED}
\textbf{Ours}                                      & 2.75  & 62.09  & 19.38 \\
\bottomrule
\end{tabular}}
\end{table}

As shown in Table~\ref{tab:complexity_runtime}, TCA-Net maintains a compact parameter size of 2.75M, which is substantially smaller than those of LLFlow and LLFormer. Although the proposed frequency-domain interaction and intensity-chromaticity fusion introduce additional FLOPs, the inference time remains lower than those of LLFlow and LLFormer. Compared with lightweight methods such as CIDNet and RetinexFormer, TCA-Net requires a moderate increase in computational cost but achieves stronger restoration performance, as shown in Table~\ref{tab:performance_comparison}. At the selection-rule level, TCA forms the sparse mask through element-wise threshold comparison followed by re-normalization, whereas Top-$\mathrm{K}$ sparsification requires ranking or sorting attention scores to enforce a fixed quota. Therefore, the extra cost of TCA-Net mainly comes from feature restoration modules, while the proposed pruning rule itself avoids explicit rank-based sorting overhead. These results suggest that TCA-Net provides a favorable trade-off among restoration quality, model compactness, and inference efficiency, rather than simply minimizing theoretical complexity.

\begin{figure*}[!htbp]
    \centering
    \includegraphics[width=\linewidth]{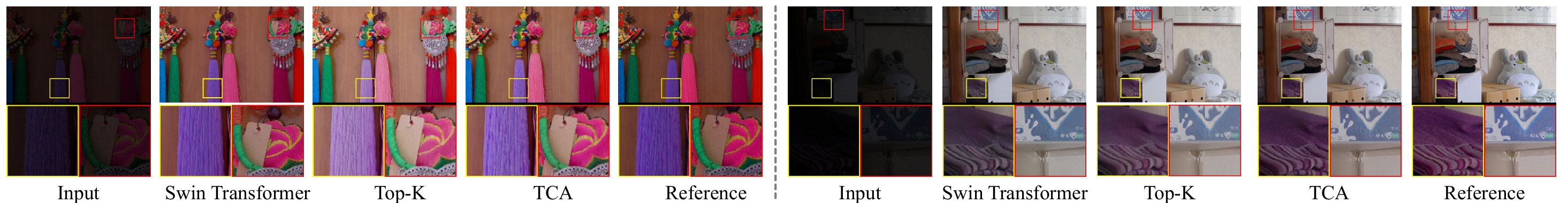}
    \caption{Qualitative comparison of different attention mechanisms on the LOL-v1 dataset. The enlarged yellow and red regions highlight local texture and color restoration details. 
    }
    \label{fig:179}
\end{figure*}

\begin{table*}[!htbp]
\centering
\caption{Sensitivity analysis of the threshold $\tau$ in TCA on different LOL benchmarks.}
\label{tab:ablation_tca_threshold}
\setlength{\tabcolsep}{4pt}
\begin{tabular}{c|ccc|ccc|ccc}
\toprule
\multirow{2}{*}{Threshold} 
& \multicolumn{3}{c|}{LOL-v1} 
& \multicolumn{3}{c|}{LOL-v2-real} 
& \multicolumn{3}{c}{LOL-v2-synthetic} \\
\cmidrule(lr){2-4} \cmidrule(lr){5-7} \cmidrule(lr){8-10}
& PSNR$\uparrow$ & SSIM$\uparrow$ & LPIPS$\downarrow$
& PSNR$\uparrow$ & SSIM$\uparrow$ & LPIPS$\downarrow$
& PSNR$\uparrow$ & SSIM$\uparrow$ & LPIPS$\downarrow$ \\
\midrule
0     
& 25.016 & 0.871 & 0.076 
& 23.390 & 0.892 & 0.118
& 25.706 & 0.942 & 0.042 \\

0.020 
& 24.589 & 0.874 & 0.072 
& 23.564 & 0.870& 0.124 
& 25.669 & 0.941 & 0.044 \\

\rowcolor[HTML]{EDEDED}
0.025 
& \textbf{25.322} & 0.876 & 0.069 
& \textbf{24.171} & 0.872  &\textbf{0.109} 
& \textbf{26.023} & \textbf{0.944} & \textbf{0.040} \\

0.030 
& 24.781 & 0.869 & \textbf{0.066} 
& 23.190 & 0.862 & 0.128
& 25.776 & 0.942 & 0.042 \\

0.040 
& 25.167 & \textbf{0.880} & 0.071 
& 23.555 &  \textbf{0.876} &  0.111
& 25.463 & 0.928 & 0.047 \\
\bottomrule
\end{tabular}
\end{table*}

\subsection{Ablation Study}

We conduct a series of ablation experiments on the LOL-v1 dataset to verify the effectiveness of the proposed components.

\textbf{Component ablation.}
To analyze the contribution of PFIM, DDSGM, and SACR, we evaluate different model variants on the LOL-v1 dataset. Unless otherwise specified, TCA is used with $\tau=0.025$. As shown in Table~\ref{tab:ablation1}, each component contributes from a different perspective. PFIM alone achieves a PSNR of 23.792 dB, indicating that phase-guided frequency interaction provides useful structural cues for illumination recovery. DDSGM and SACR alone obtain higher PSNR values of 24.458 dB and 24.392 dB, respectively, demonstrating the benefits of residual intensity-guided fusion and scale-aware consistency regularization. When combined, these modules show clear complementary effects. In particular, PFIM+DDSGM improves the PSNR to 25.128 dB, and the full model obtains the strongest PSNR and LPIPS among the variants while maintaining competitive SSIM. These results verify that the proposed modules collaboratively improve illumination recovery, color fidelity, and perceptual quality.

\begin{table}[!htb] 
\centering 
\caption{Ablation study on the contribution of PFIM, DDSGM, and SACR on the LOL-v1 dataset.} \label{tab:ablation1} 
\begin{tabular}{ccc|ccc} 
\rowcolor[HTML]{EDEDED}
\toprule PFIM & DDSGM & SACR & PSNR$\uparrow$ & SSIM$\uparrow$ & LPIPS$\downarrow$ \\ \midrule 
\xmark & \xmark & \xmark & 23.809 & 0.857 & 0.079 \\ 
\cmark & \xmark & \xmark & 24.099 & 0.852 & 0.113 \\ 
\xmark & \cmark & \xmark & 24.458 & \textbf{0.878} & 0.078 \\ 
\xmark & \xmark & \cmark & 24.392 & 0.860 & 0.080 \\ 
\cmark & \cmark & \xmark & 25.128 & 0.876 & 0.072 \\ 
\cmark & \xmark & \cmark & 24.300 & 0.858 & 0.095 \\ 
\xmark & \cmark & \cmark & 24.570 & 0.872 & 0.077 \\ 

\rowcolor[HTML]{EDEDED}
\cmark & \cmark & \cmark & \textbf{25.322} & {0.876} & \textbf{0.069} \\ 
\bottomrule 
\end{tabular} 
\end{table}

\textbf{Analysis of the threshold in TCA.}
We further analyze the influence of the confidence threshold $\tau$ in TCA. This threshold controls the pruning strength of low-confidence cross-stream attention responses after Softmax normalization, so it is defined on a probability-level confidence scale rather than on dataset-dependent feature magnitudes. When $\tau=0$, TCA degenerates into dense cross-attention without confidence-based pruning. As reported in Table~\ref{tab:ablation_tca_threshold}, we evaluate five threshold values, i.e., 0, 0.020, 0.025, 0.030, and 0.040. Compared with dense cross-attention, setting $\tau=0.025$ improves the PSNR from 25.016 dB to 25.322 dB and also improves SSIM from 0.871 to 0.876, indicating that removing weak attention responses helps suppress unreliable cross-stream interactions. Although $\tau=0.030$ achieves the lowest LPIPS and $\tau=0.040$ obtains the highest SSIM, both settings lead to lower PSNR, suggesting that overly aggressive pruning may discard useful dependencies for pixel-wise restoration. Therefore, we choose $\tau=0.025$ as the final setting because it provides the strongest PSNR while maintaining competitive SSIM and LPIPS across the evaluated LOL benchmarks. Although TCA yields input- and layer-dependent sparse patterns through a fixed probability-level threshold, the threshold value itself is shared across inputs during inference. This simple design keeps the pruning rule stable and efficient, but a more flexible image- or layer-conditioned threshold may further adapt the pruning strength to varying illumination conditions, noise levels, and cross-stream attention distributions. We leave adaptive threshold estimation as future work.

\begin{table}[!htbp]
    \centering
    \caption{Ablation study on different sparse attention mechanisms.}
    \label{tab:performance_comparison_topk}
    \begin{tabular}{lccc}
        \toprule
        \rowcolor[HTML]{EDEDED}
        Method & PSNR$\uparrow$ & SSIM$\uparrow$ & LPIPS$\downarrow$\\
        \midrule
        Full Cross-Attention w/o Pruning & 25.016 & 0.871 & 0.076\\
        Window-based Sparse Attention & 24.222 & 0.866 & 0.072 \\
        Top-$\mathrm{K}$ Attention ($\rho=0.6$) & 25.093 & 0.871 &0.070 \\
        Top-$\mathrm{K}$ Attention ($\rho=0.7$) & 24.561 & 0.869 &0.073\\
        \rowcolor[HTML]{EDEDED}
        TCA (Ours) & \bf 25.322 & \bf 0.876 & \bf0.069\\
        \bottomrule
    \end{tabular}
\end{table}


\textbf{Ablation on attention selection rules.}
To further validate the effectiveness of TCA, we compare it with representative attention variants on the LOL-v1 dataset, including dense cross-attention, window-based sparse attention, and Top-$\mathrm{K}$ sparse attention~\cite{chen2023learning}. Dense cross-attention retains all cross-stream responses, window-based attention restricts interactions to fixed local regions, and Top-$\mathrm{K}$ attention keeps a fixed proportion of the highest-ranking responses. In contrast, TCA removes low-confidence cross-stream dependencies according to Softmax-normalized attention probabilities.

As shown in Table~\ref{tab:performance_comparison_topk}, dense cross-attention achieves a PSNR of 25.016 dB, but it may propagate noise-corrupted responses because all non-zero attention weights are retained. Window-based sparse attention obtains a lower PSNR of 24.222 dB, suggesting that fixed local windows may limit global illumination recovery. 
For Top-$\mathrm{K}$ attention, the two predefined fixed retention ratios $\rho=0.6$ and $\rho=0.7$ yield PSNR values of 25.093 dB and 24.561 dB, respectively. Neither setting achieves optimal model performance, as fixed pruning quotas ignore layer-wise and input-specific variations in attention distributions.
By comparison, TCA obtains the strongest PSNR and SSIM in this comparison, with 25.322 dB PSNR and 0.876 SSIM. These results show that confidence-based pruning is more effective for reliable intensity-chromaticity fusion than dense interaction, fixed local windows, or fixed-ratio selection.

\textbf{Visual analysis with representative attention mechanisms.}
To further validate TCA, we compare it with Top-$\mathrm{K}$ sparsification and Swin Transformer attention on the LOL-v1 dataset~\cite{wei2018deep}, as shown in Fig.~\ref{fig:179}. In the upper row, Top-$\mathrm{K}$ with $\rho=0.7$ introduces visible color shifts and distortions in the zoomed-in regions, suggesting that its fixed retention ratio may preserve noisy responses while discarding useful dependencies. In the lower row, Swin Transformer attention improves brightness but still produces blurred textures and desaturated colors, indicating that fixed local-window attention may limit long-range illumination and texture recovery.

In contrast, TCA produces results closer to the reference in both cases. By pruning low-confidence cross-stream interactions according to Softmax-normalized attention probabilities, TCA better preserves informative dependencies, leading to more faithful color restoration and cleaner structural details.

\section{Conclusion}

This paper presents TCA-Net for reliable intensity-chromaticity fusion in the HVI space. TCA-Net improves cross-stream fusion through three coordinated designs: PFIM provides phase-guided frequency initialization for the intensity stream, TCA suppresses low-confidence interactions using a fixed confidence threshold with input-adaptive retained cardinality, and DDSGM reduces chromaticity-related interference via residual intensity guidance. SACR further regularizes the restored output with a decayed scale-consistency constraint during training. Experiments on five LLIE benchmarks show that TCA-Net achieves competitive performance on the LOL datasets, improves chromaticity fidelity on LOL-v1, and remains effective on extremely dark and mobile low-light scenarios while using only 2.75M parameters.

\bibliographystyle{IEEEtran}
\bibliography{reference}

\end{document}